\documentclass[11pt]{article} 

\usepackage{iftex}
\ifpdf
  \usepackage{graphicx}
  \usepackage{xcolor}
  \usepackage{microtype}
\else
  \usepackage[dvipdfmx]{graphicx}
  \usepackage[dvipdfmx]{xcolor}
  \usepackage[DVIpsnames, dvipdfmx]{microtype}
\fi

\usepackage{url}
\usepackage{cite}
\usepackage{amsmath,amssymb,amsfonts}

\usepackage{authblk}

\usepackage{algorithm}
\usepackage{algpseudocode}

\usepackage{booktabs}
\usepackage{textcomp}
\usepackage{enumitem}
\usepackage{pgfplotstable}
\usepackage{pgfplots}
\pgfplotsset{compat=newest} 

\def\BibTeX{{\rm B\kern-.05em{\sc i\kern-.025em b}\kern-.08em
    T\kern-.1667em\lower.7ex\hbox{E}\kern-.125emX}}

\title{Online Adaptive Reinforcement Learning with Echo State Networks for Non-Stationary Dynamics \\

}
\author[1]{Aoi Yoshimura}
\author[1,2]{Gouhei Tanaka}

\affil[1]{Department of Computer Science, Nagoya Institute of Technology, \authorcr Nagoya 466-8555, Japan}

\affil[2]{International Research Center for Neurointelligence, \authorcr The University of Tokyo, Tokyo 113-0033, Japan}

\affil[ ]{\texttt{a.yoshimura.536@stn.nitech.ac.jp, gtanaka@nitech.ac.jp}}

\date{} 

\begin{document}
\maketitle

\begin{abstract}
Reinforcement learning (RL) policies trained in simulation often suffer from severe performance degradation when deployed in real-world environments due to non-stationary dynamics. While Domain Randomization (DR) and meta-RL have been proposed to address this issue, they typically rely on extensive pretraining, privileged information, or high computational cost, limiting their applicability to real-time and edge systems.
In this paper, we propose a lightweight online adaptation framework for RL based on Reservoir Computing. Specifically, we integrate an Echo State Networks (ESNs) as an adaptation module that encodes recent observation histories into a latent context representation, and update its readout weights online using Recursive Least Squares (RLS). This design enables rapid adaptation without backpropagation, pretraining, or access to privileged information.
We evaluate the proposed method on CartPole and HalfCheetah tasks with severe and abrupt environment changes, including periodic external disturbances and extreme friction variations. Experimental results demonstrate that the proposed approach significantly outperforms DR and representative adaptive baselines under out-of-distribution dynamics, achieving stable adaptation within a few control steps. Notably, the method successfully handles intra-episode environment changes without resetting the policy. Due to its computational efficiency and stability, the proposed framework provides a practical solution for online adaptation in non-stationary environments and is well suited for real-world robotic control and edge deployment.
\end{abstract}

\vspace{1ex} 

\section{Introduction}

Deep Reinforcement Learning (DRL) has demonstrated remarkable success in high-dimensional control tasks, ranging from game playing to complex robotic manipulation. Algorithms such as Proximal Policy Optimization (PPO)~\cite{schulman2017proximal} and Soft Actor-Critic (SAC)~\cite{haarnoja2018softactorcriticoffpolicymaximum} enable agents to learn optimal control policies in an end-to-end manner. However, deploying these policies to real-world robotic systems remains a significant challenge due to the well-known \emph{Sim2Real gap}. Simulators rely on idealized physical models, and inevitable discrepancies in parameters such as friction coefficients, contact dynamics, and sensor noise often lead to severe performance degradation when policies trained in simulation are transferred to the real world~\cite{tobin2017domain,peng2018sim}.

Moreover, real-world environments are inherently non-stationary. Physical properties can change abruptly during manipulation, control, and decision-making tasks due to external disturbances and friction variations. Such changes may occur even within a single episode during RL, rendering static policies with fixed parameters insufficient. Addressing both partial observability and intra-episode environmental variation is therefore essential for robust real-world deployment.

To mitigate the Sim2Real gap, Domain Randomization (DR) has become a widely adopted technique, where physical parameters are randomized during training to improve robustness~\cite{tobin2017domain,peng2018sim}. While effective for disturbances within the training distribution, DR fundamentally relies on interpolation and struggles under out-of-distribution (OOD) dynamics, often resulting in overly conservative policies that sacrifice optimality for stability~\cite{dulac2019challenges}. These limitations motivate the development of explicit online adaptation mechanisms.

Recent work has explored history-based adaptation methods that infer latent environmental parameters from observation sequences. A representative example is Rapid Motor Adaptation (RMA)~\cite{kumar2021rmarapidmotoradaptation}, which estimates dynamics-related context variables using a learned encoder. Despite its effectiveness, RMA requires extensive offline pretraining with privileged information, such as ground-truth physical parameters, and large-scale simulation data. Such requirements significantly increase development cost and limit scalability. Alternatively, test-time training approaches, including Policy Adaptation during Deployment (PAD), perform online adaptation using self-supervised learning objectives. However, these methods rely on backpropagation during deployment, leading to high computational overhead and potential instability, which are undesirable for resource-constrained edge devices requiring real-time control.

In this paper, we propose a lightweight real-time adaptation framework, termed \emph{ESN-based Online Adaptation  (ESN-OA), in the context of reinforcement learning}. Unlike existing meta-learning or test-time training approaches, our method requires no pretraining phase for adaptation and no access to privileged information. We integrate an Echo State Network (ESN)~\cite{jaeger2001echo,LUKOSEVICIUS2009127} into a DRL agent to extract temporal context from recent observation histories. The key idea is to train only the ESN readout layer online using Recursive Least Squares (RLS). This approach avoids the computationally expensive backpropagation required for updating deep neural network weights, enabling rapid and stable adaptation with minimal overhead.

The main contributions of this paper are summarized as follows:
\begin{enumerate}
    \item \textbf{Zero-shot Adaptation to Dynamics Shift:} We propose an online adaptation framework that enables agents to adapt to non-stationary environments from scratch during deployment. We refer to this framework as zero-shot adaptation to dynamics shift, since the adaptation module requires no prior exposure to target dynamics variations (meta-training) or privileged information.
    
    \item \textbf{Computational Efficiency for Edge Deployment:} By leveraging RLS for online readout updates, the proposed method achieves fast convergence and low computational overhead, making it suitable for real-time robotic control on resource-constrained platforms.
    \item \textbf{Robustness to Intra-Episode Dynamics Changes:} We demonstrate that the proposed approach can handle sudden and extreme changes in physical parameters, such as multi-fold variations in friction, within a few control steps, significantly outperforming DR and representative adaptive baselines.
\end{enumerate}

\section{Related Work}

\subsection{Domain Randomization and its Limits}
While DR is the standard approach for bridging the Sim2Real gap~\cite{da2025surveysimtorealmethodsrl,muratore2022robot}, standard DR relies on fixed parameter distributions designed manually~\cite{mozian2020learning}. Recent advancements include Active DR~\cite{pmlr-v100-mehta20a} and entropy-maximization techniques~\cite{tiboni2024domainrandomizationentropymaximization} to improve sample efficiency and coverage. Alternatively, recent work proposes bridging this gap by constructing high-fidelity simulators that incorporate realistic physics such as friction and air resistance~\cite{10602859}. However, these methods result in static policies that struggle to generalize to OOD scenarios~\cite{zhang2024mixturedatatrainingensure,curtis2025flowbaseddomainrandomizationlearning}. As noted in recent surveys~\cite{malmir2025diarelreinforcementlearningdisturbance,ditzler2015learning}, static policies remain insufficient for non-stationary real-world environments, necessitating mechanisms that can adapt to environmental changes on the fly.

\subsection{POMDPs and Meta-Reinforcement Learning}
To handle partial observability and dynamics shifts, control tasks are often formulated as Partially Observable Markov Decision Processes (POMDPs) requiring history-based inference~\cite{ponnambalam2022back,kim2023reference}. 
Transformer-based architectures have been employed to capture long-term context~\cite{lu2024rethinkingtransformerssolvingpomdps,vaswani2017attention}, and Meta-Rein\-force\-ment Learning (Meta-RL) approaches like RMA~\cite{kumar2021rmarapidmotoradaptation} demonstrate robust adaptation by estimating environmental extrinsics. Similarly, factored adaptation frameworks have been proposed to explicitly model time-varying latent factors in non-stationary environments~\cite{feng2022factored}. However, these methods typically require extensive pre-training with privileged information (e.g., ground-truth physics) which is unavailable in many real-world settings. Alternatively, test-time training methods such as PAD~\cite{hansen2021selfsupervisedpolicyadaptationdeployment} adapt without privileged information but rely on backpropagation during deployment. This incurs high computational cost and latency~\cite{yin2023accurate}, making them ill-suited for resource-constrained edge devices compared to the gradient-free adaptation proposed in this work.

\section{Proposed Method: Online Adaptive RL using ESN}
\label{sec:method}

\begin{figure*}[t]
    \centering
    \includegraphics[width=1.0\textwidth]{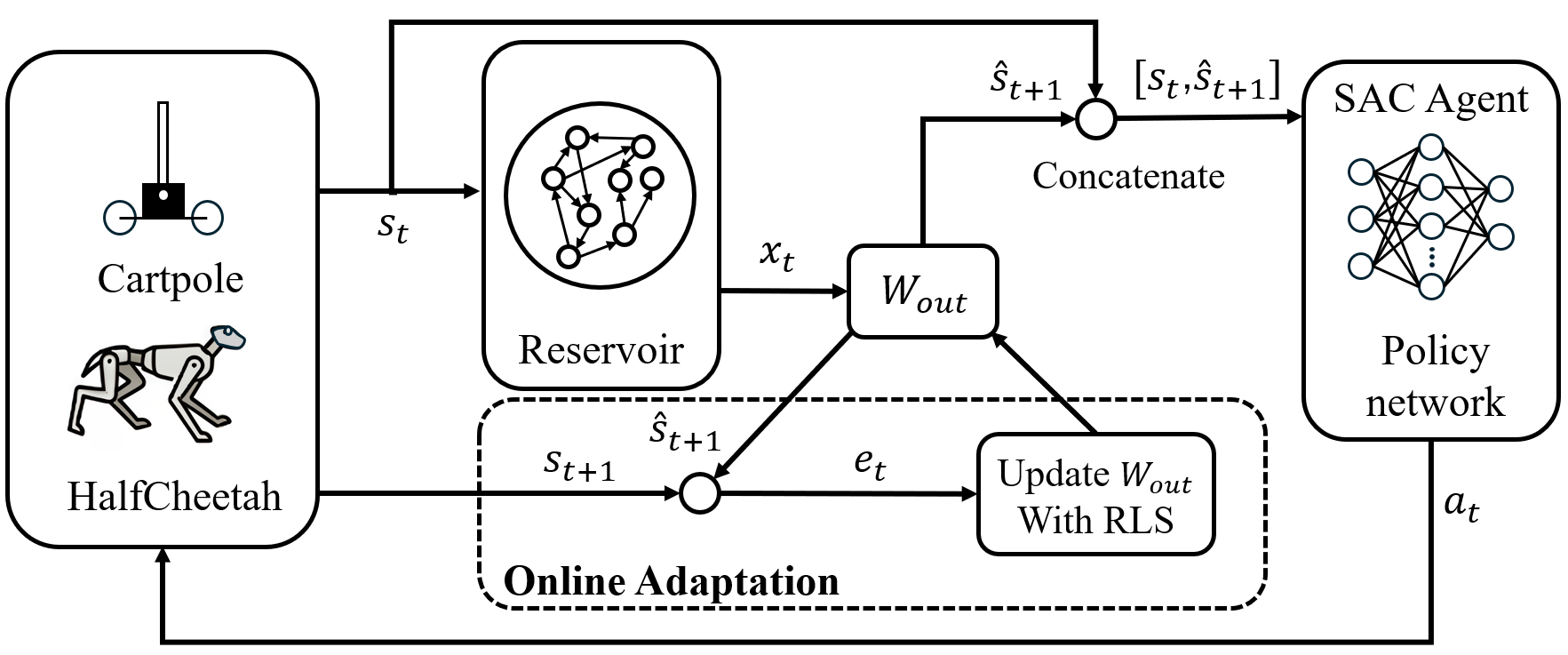}
    \caption{Overview of the proposed ESN-based online adaptation system. The reservoir captures temporal context, and RLS updates the readout weights $W^{\mathrm{out}}$ in real-time based on prediction error $e_t$. The predicted next state $\hat{s}_{t+1}$ is concatenated with the current state $s_t$ and used as the input to the policy network of the SAC agent.}
    \label{fig:system_overview}
\end{figure*}

In real-world robotic control, the assumption of stationarity often fails due to external disturbances and friction variations. While DR and Meta-RL attempt to address this issue as described in Section~II, they typically require extensive pre-training or privileged information. To overcome these limitations, we propose an \textit{Online Adaptive Reinforcement Learning} framework based on Reservoir Computing, specifically utilizing an ESN. Our approach utilizes the RLS-based online learning for the readout layer, enabling rapid adaptation to non-stationary dynamics without any pre-training. The overall system architecture is illustrated in Fig.~\ref{fig:system_overview}, and the pseudo-code of the learning procedure is presented in Algorithm \ref{alg:esn_oa}.

\subsection {Echo State Networks}
Recurrent Neural Networks (RNNs) and their variants widely used for processing time-series data and estimating hidden states. However, despite advances of RNN variants such as LSTM and GRU, training RNN-based models via Backpropagation Through Time (BPTT) remains computationally expensive and can still suffer from vanishing gradient issues in long-sequence modeling. To address this, we employ Reservoir Computing (RC), specifically ESN architecture.

The ESN consists of a fixed, randomly initialized recurrent layer (reservoir) and a trainable output layer (readout). Let $u_t \in \mathbb{R}^{N_u}$ be the input vector, $x_t \in \mathbb{R}^{N_x}$ be the internal reservoir state, and $y_t \in \mathbb{R}^{N_y}$ be the output vector at time step $t$. The state update rule is defined as follows\cite{jaeger2001echo,JAEGER2007335}:
\begin{equation}
    x_t = (1 - \alpha)x_{t-1} + \alpha \cdot \tanh(W^{\mathrm{in}}u_t + W^{\mathrm{res}}x_{t-1}),
    \label{eq:esn_update}
\end{equation}
where $W^{\mathrm{in}} \in \mathbb{R}^{N_x \times N_u}$ is the input weight matrix, $W^{\mathrm{res}} \in \mathbb{R}^{N_x \times N_x}$ is the recurrent weight matrix, and $\alpha \in (0, 1]$ is the leak rate. To ensure the Echo State Property, the spectral radius of $W^{\mathrm{res}}$ is set to $\rho(W^{\mathrm{res}}) < 1$ \cite{LUKOSEVICIUS2009127}.

The network output $\hat{y}_t$ is computed as a linear combination of the reservoir states:
\begin{equation}
    \hat{y}_t = W^{\mathrm{out}}x_t,
    \label{eq:readout}
\end{equation}
where $W^{\mathrm{out}} \in \mathbb{R}^{N_y \times N_x}$ is the only trainable weight matrix in our adaptation module.

\subsection{Online Adaptation via Recursive Least Squares}
To enable rapid online adaptation, we employ the RLS algorithm for updating $W^{\mathrm{out}}$. RLS provides second-order convergence properties, 
which allows for immediate adaptation to non-stationary environments within a few time steps \cite{haykin2014adaptive}.

The objective is to minimize the weighted sum of squared errors between the predicted next state $\hat{s}_{t+1}$ and the actual next state $s_{t+1}$, given by:
\begin{equation}
    \mathcal{L}_t = \sum_{i=1}^{t} \lambda^{t-i} \| e_i \|^2, 
\end{equation}
\begin{equation}
    e_i = s_{i+1} - W^{\mathrm{out}}x_i.
    \label{eq:e(i)}
\end{equation}
Here, $\lambda \in (0, 1]$ is the forgetting factor. A smaller $\lambda$ allows the system to rapidly forget past dynamics and adapt to sudden environmental changes.

The online update procedure for $W^{\mathrm{out}}$ at each step $t$ is as follows:
\begin{enumerate}
    \item Compute Prediction Error:
    \begin{equation}
       e_t = s_{t+1} - W^{\mathrm{out}}_{t-1}x_t.
       \label{eq:e(t)}
    \end{equation}
    \item Compute Gain Vector:
    \begin{equation}
       k_t = \frac{P_{t-1}x_t}{\lambda + x_t^T P_{t-1}x_t},
       \label{eq:k(t)}
   \end{equation}
   where the precision matrix $P_t$ approximates the inverse covariance matrix $(\sum_t x_t x_t^T)^{-1}$.
    \item Update Output Weights:
    \begin{equation}
       W^{\mathrm{out}}_t \leftarrow W^{\mathrm{out}}_{t-1} + e_tk_t^T.
       \label{eq:W(out)t}
    \end{equation}
     \item Update Inverse Covariance Matrix:
     \begin{equation}
       P_t = \frac{1}{\lambda} (P_{t-1} - k_tx_t^T P_{t-1}).
       \label{eq:P(t)}
   \end{equation}
\end{enumerate}
Crucially, for the ``No pre-training'' setting, we initialize $W^{\mathrm{out}}_0 = 0$ and $P_0 = \delta I$ (where $\delta$ is a large constant, e.g., 100). This setting of $P_0$ reflects a state of high sensitivity to prediction errors, facilitating rapid initial adaptation of $W^{\mathrm{out}}$.

\subsection{Integration with Soft Actor-Critic (SAC)}

\begin{algorithm}[t!]
\caption{ESN-based Online Adaptation (ESN-OA)}
\label{alg:esn_oa}
\begin{algorithmic}[1]
\State \textbf{Input:} Reservoir ($N_x, \rho, \alpha$), RLS ($\lambda, \delta$), Policy $\pi_\phi$
\State \textbf{Init:} $W^{\mathrm{in}}, W^{\mathrm{res}}$ (random); $W^{\mathrm{out}} \leftarrow \mathbf{0}$, $P \leftarrow \delta I$, $x_0 \leftarrow \mathbf{0}$

\For{each episode}
    \State Reset environment and observe $s_0$
    \Statex \hspace{\algorithmicindent}(Nominal for training, Non-stationary for testing)
       \For{$t = 0$ to $T$}
        \State Update reservoir state: $x_t \leftarrow (1 - \alpha)x_{t-1} +$
        \Statex \hspace{\algorithmicindent}\hspace{\algorithmicindent}$\alpha \cdot \tanh(W^{\mathrm{in}}s_t + W^{\mathrm{res}}x_{t-1})$ \Comment{Eq. \eqref{eq:esn_update}}
        \State Predict next state: $\hat{s}_{t+1} \leftarrow W^{\mathrm{out}} x_t$ \Comment{Eq. \eqref{eq:readout}}
        \State Augment state: $\tilde{s}_t \leftarrow [s_t, \hat{s}_{t+1}]$ \Comment{Eq. \eqref{eq:s(t)}}
        \State Select action $a_t \sim \pi_\phi(\cdot | \tilde{s}_t)$ 
        \State Execute $a_t$, observe $r_t, s_{t+1}$
        
        \State \textit{// Online Adaptation (RLS)}
        \State $e \leftarrow s_{t+1} - \hat{s}_{t+1}$ \Comment{Eq. \eqref{eq:e(t)}}
        \State $k \leftarrow \frac{P x_t}{\lambda + x_t^T P x_t}$ \Comment{Eq. \eqref{eq:k(t)}}
        \State $W^{\mathrm{out}} \leftarrow W^{\mathrm{out}} + e k^T$ \Comment{Eq. \eqref{eq:W(out)t}}
        \State $P \leftarrow \frac{1}{\lambda} (P - k x_t^T P)$ \Comment{Eq. \eqref{eq:P(t)}}
        \If
        {training}
        \State  Update SAC parameters $\phi$ using standard loss
        \EndIf
    \EndFor
\EndFor
\end{algorithmic}
\end{algorithm}

We integrate the ESN-RLS adaptation module with a SAC agent. 

The standard policy $\pi(a_t|s_t)$ in MDPs relies only on the current observation $s_t$. However, under partial observability caused by hidden physical parameters (e.g., friction coefficients, wind force), $s_t$ is insufficient for adaptation to non-stationary environments. To address this, our system operates as follows:

\begin{enumerate}
    \item Context extraction: The reservoir receives the current observation $s_t$ and updates its internal state $x_t$, encoding the temporal history of the trajectory.
    \item Dynamics prediction: The readout layer predicts the next state $\hat{s}_{t+1} = W^{\mathrm{out}}x_t$. Since $W^{\mathrm{out}}$ is updated online via RLS to minimize prediction error, $\hat{s}_{t+1}$ implicitly contains information about the current environmental dynamics.
    \item State augmentation: The policy network receives an augmented state vector $\tilde{s}_t$ formed by concatenating the raw observation and the ESN prediction:
    \begin{equation}
        \tilde{s}_t = [s_t, \hat{s}_{t+1}].
        \label{eq:s(t)}
    \end{equation}
    \item Action generation: The SAC agent outputs action $a_t = \pi(\tilde{s}_t)$ based on this context-aware state.
\end{enumerate}

By using the predicted next state $\hat{s}_{t+1}$ instead of the high-dimensional reservoir state $x_t$ for concatenation, we avoid the curse of dimensionality while providing the agent with explicit information about the estimated dynamics.

\section{Experiments}
To validate the efficacy of the proposed ESN-based online adaptable RL framework, we conducted comparative experiments in two simulated environments: CartPole \cite{Barto1983NeuronlikeAE} and HalfCheetah \cite{wawrzynski2009cat}. The primary objective is to evaluate the system's capability for zero-shot adaptation, such as maintaining control performance in non-stationary environments without any pre-training or prior data collection.

\subsection{Experimental Environments and Non-stationarity}
We introduced specific non-stationary conditions into the physical parameters of the environments to simulate external disturbances and sudden hardware failures.

\begin{enumerate}
    \item CartPole with periodic disturbance
    
    The CartPole task requires an agent to balance a pole vertically by applying horizontal forces to a cart. The state space consists of the cart position, cart velocity, pole angle, and pole angular velocity.
    The standard CartPole environment ($g=9.8~\mathrm{m/s^2}$ , cart mass $1.0~\mathrm{kg}$, pole mass $0.1~\mathrm{kg}$) was utilized \cite{towers2025gymnasiumstandardinterfacereinforcement}. In addition, a time-varying external wind force $F _{\mathrm{wind}}=A\cos ( \omega t)$ was applied to the cart. 
\end{enumerate}

\begin{itemize} 
    \item \textbf{Protocol:} During training, the environment was stationary ($A=0$). During testing, the amplitude $A$ was varied within the range $[1.0,10.0]$ to assess robustness against non-stationary periodic disturbances. 
\end{itemize}

\begin{enumerate}[resume]
    \item HalfCheetah with a sudden parameter change 
    
    HalfCheetah is a high-dimensional (17-dimensional) locomotion task where a planar cheetah-like robot learns to run forward as fast as possible. The agent must coordinate the torques of six joints based on the positions and velocities of its body parts.
    For the HalfCheetah task, we simulated a drastic change in contact dynamics to evaluate the temporal adaptation speed.
\end{enumerate}

\begin{itemize} 
    \item \textbf{Protocol:} The friction coefficient of the ground was maintained at its default value ($1.0\times$) for the first 500 time steps. At step 500, the coefficient was abruptly scaled by a factor ranging from $1.0\times$ to $10.0\times$. This setup tests the agent's ability to recover from sudden ``failures'' in real-time. 
\end{itemize}

\subsection{Baselines and Comparative Analysis}
We compared the proposed method against six baseline strategies, including state-of-the-art domain adaptation methods (PAD and RMA). A critical aspect of this comparison is the data collection strategy required for pre-training based methods.

\begin{enumerate}
    \item Proposed Method
\end{enumerate} 
\begin{itemize} 
    \item \textbf{ESN-OA (ESN-based Online Adaptation without pre-training):} The readout weights $W^{\mathrm{out}}$ are initialized to zero, and adaptation via RLS begins immediately at the start of the test episode. This method requires absolutely no pre-training phase. 
\end{itemize}
\begin{enumerate}[resume]
    \item Pre-trained Baselines
\end{enumerate} 
\begin{itemize} 
    \item \textbf{ESN-OA-PT (ESN-based Online Adaptation with pre-training):} The weights $W^{\mathrm{out}}$ are pre-trained on a stationary environment. To ensure high-quality reservoir feature extraction, the pre-training data must contain long, continuous trajectories. Since purely random actions lead to early termination in CartPole, we employed a ``noisy rule-based policy'' (30~\% random actions) to secure trajectory length while maintaining exploration. Similarly, for HalfCheetah, we utilized Ornstein-Uhlenbeck (OU) noise ($\omega=0.3$) to generate temporally correlated motion data suitable for reservoir dynamics learning.
    \item \textbf{PAD (Policy Adaptation during Deployment):} A self-supervised adaptation method that trains an encoder using an auxiliary loss $\mathcal{L}_{\mathrm{total}} = \mathcal{L}_{\mathrm{IDM}} + \mathcal{L}_{\mathrm{rec}}$ (Inverse Dynamics Model + Reconstruction) ~\cite{hansen2021selfsupervisedpolicyadaptationdeployment}.
    In contrast to the ESN approach, PAD relies on learning a diverse latent representation of the environment rather than capturing long temporal dependencies. Consequently, to prevent overfitting and ensure broad state-space coverage during pre-training, we prioritized diversity over stability by utilizing a high-entropy policy ($50~\%$ random mixture) for CartPole and DR for HalfCheetah.
    \item \textbf{RMA (Rapid Motor Adaptation):} A two-phase method using an oracle policy and a history-based adaptor network (50-step history)~\cite{kumar2021rapid}.
\end{itemize}
\begin{enumerate}[resume]
    \item Standard Baselines
\end{enumerate}
\begin{itemize}
    \item \textbf{Standard SAC:} A standard SAC agent trained in a stationary environment, used as a lower-bound reference to highlight the performance degradation under distribution shift.
    \item \textbf{DR:} A robust policy trained with randomized physical parameters, serving as a baseline for comparison against static robustness versus explicit online adaptation.
\end{itemize}

\subsection{Implementation Details}
All algorithms were implemented using Python libraries, PyTorch and Stable Baselines3 \cite{JMLR:v22:20-1364}. Hyperparameters were selected via grid search.
Experiments were conducted on a machine with an AMD Ryzen 9 9950X CPU and 128 GB RAM. To ensure statistical reliability, all experiments for each method were conducted over $N=10$ independent trials with different random seeds.
\subsubsection{Reinforcement Learning (SAC)} The SAC agent utilized an Multilayer Perceptron (MLP) architecture consisting of two layers with hidden sizes of $[64, 64]$ for CartPole and $[256, 256]$ for HalfCheetah. The optimizer was Adam \cite{kingma2014adam} with a learning rate of $3\times10^{-4}$.
\subsubsection{Reservoir and RLS Settings} The ESN configuration was tailored to the task complexity: 
\begin{itemize} 
    \item \textbf{Reservoir Size ($N_x$):} 300 for CartPole, 500 for HalfCheetah. 
    \item \textbf{Dynamics:} Spectral radius $\rho=0.9$, Leak rate $\alpha=0.3$. 
    \item \textbf{Online Adaptation (RLS):} The forgetting factor $\lambda$ was set to 0.99 for CartPole and 0.95 for HalfCheetah to handle the rapid dynamics changes. The inverse covariance matrix was initialized as $P_0=100I$ to facilitate rapid initial learning. 
\end{itemize}

\section{Results}

\subsection{Performance on CartPole with Non-Stationary Disturbances}
\begin{figure}[h]
    \centering
    \includegraphics[width=0.75\textwidth]{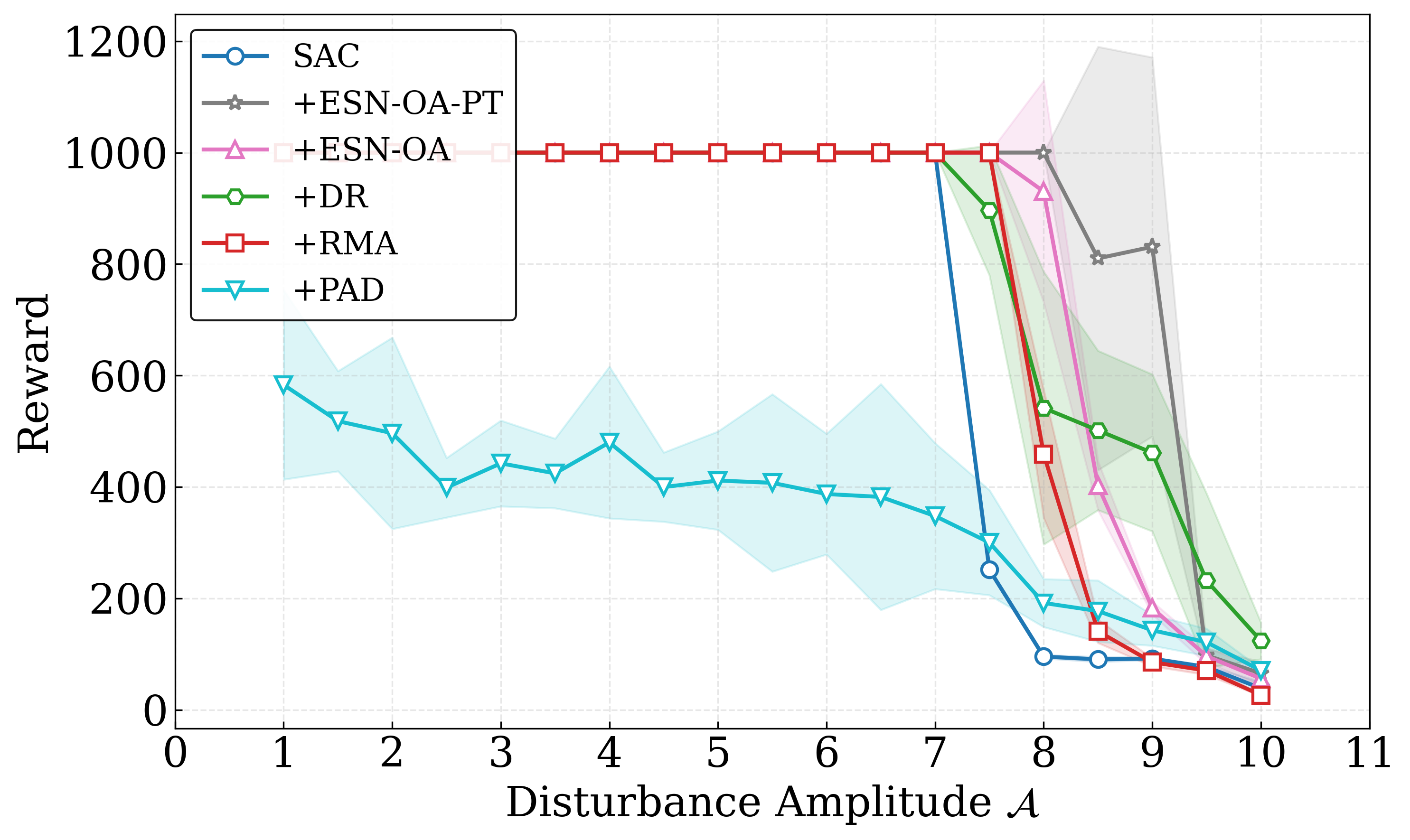}
    \caption{Average reward in the CartPole environment under increasing periodic wind disturbances.}
    \label{fig:cartpole_result}
\end{figure}

We evaluated robustness of the RL algorithms against non-stationary periodic wind disturbances in the CartPole environment. An external force $F_{\mathrm{wind}} = A \cos(\omega t)$ was applied as described in Sec. IV-A, and the disturbance amplitude $A$ was gradually increased during testing. Fig.~\ref{fig:cartpole_result} shows the average reward as a function of $A$, and Table \ref{tab:cartpole_stats} summarizes the numerical results.

\begin{table}[h]
\centering
\caption{Robustness evaluation on CartPole (mean $\pm$ std).}
\label{tab:cartpole_stats}
\resizebox{\columnwidth}{!}{%
\begin{tabular}{l|r@{ $\pm$ }lr@{ $\pm$ }lr@{ $\pm$ }lr@{ $\pm$ }l}
\hline
\textbf{Method} & \multicolumn{2}{c}{\textbf{A=8.0}} & \multicolumn{2}{c}{\textbf{A=8.5}} & \multicolumn{2}{c}{\textbf{A=9.0}} & \multicolumn{2}{c}{\textbf{A=9.5}} \\ \hline
SAC & 95.5 & 3.1 & 90.5 & 4.4 & 91.8 & 2.3 & 76.8 & 2.4 \\
+ESN-OA-PT & 1000.0 & 0.0 & 810.1 & 379.9 & 830.9 & 340.1 & 97.6 & 12.5 \\ 
+ESN-OA & 930.2 & 199.0 & 402.3 & 43.9 & 182.7 & 11.7 & 96.2 & 14.7 \\
+DR & 541.6 & 244.8 & 501.5 & 142.9 & 461.2 & 140.8 & 231.8 & 156.3 \\
+RMA & 459.0 & 114.9 & 140.9 & 20.8 & 85.4 & 7.3 & 70.4 & 7.5 \\
+PAD & 191.7 & 43.0 & 177.2 & 55.0 & 142.7 & 27.5 & 121.3 & 24.4 \\

\hline
\end{tabular}
}
\end{table}

\begin{figure}[h]
    \centering
    \includegraphics[width=0.75\textwidth]{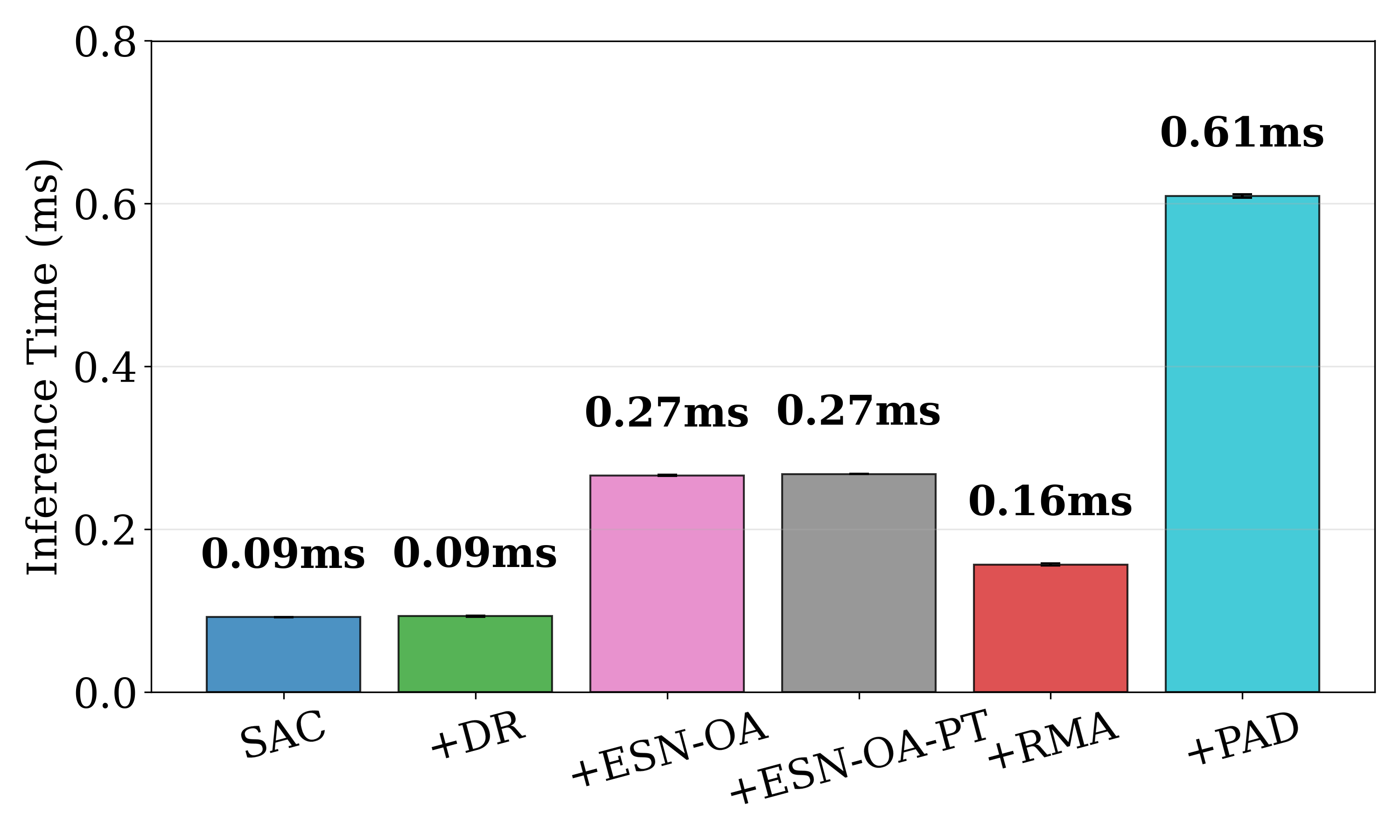}
    \caption{Inference time per control step on CartPole.}
    \label{fig:inference_time}
\end{figure}

The standard SAC and representative baselines (DR, RMA) collapse when the disturbance amplitude $A$ exceeds 8.0. In contrast, ESN-OA and ESN-OA-PT maintain near-optimal performance across the tested range ($N=10$). ESN-OA-PT shows even higher stability at extreme amplitudes ($A=9.5$), indicating that pre-training further strengthens the reservoir's representation. This robustness indicates that the latent context encoded by the ESN effectively compensates for OOD forces in real-time. Regarding computational overhead, ESN-OA achieves a practical balance between adaptation and latency. Fig.~\ref{fig:inference_time} confirms that ESN-OA requires only $0.27$~ms per step ($3.7$~kHz), significantly faster than the gradient-based PAD ($0.61$~ms). This efficiency makes it well-suited for high-frequency control.

\subsection{Performance on HalfCheetah with Sudden Parameter Shifts}
We next evaluated adaptation performance in the HalfCheetah environment under abrupt changes in ground friction. The friction coefficient was scaled by a multiplier ranging from $1.0\times$ to $10.0\times$. PAD was excluded due to unstable behavior in this task.

\begin{figure}
    \centering
    \includegraphics[width=0.75\textwidth]{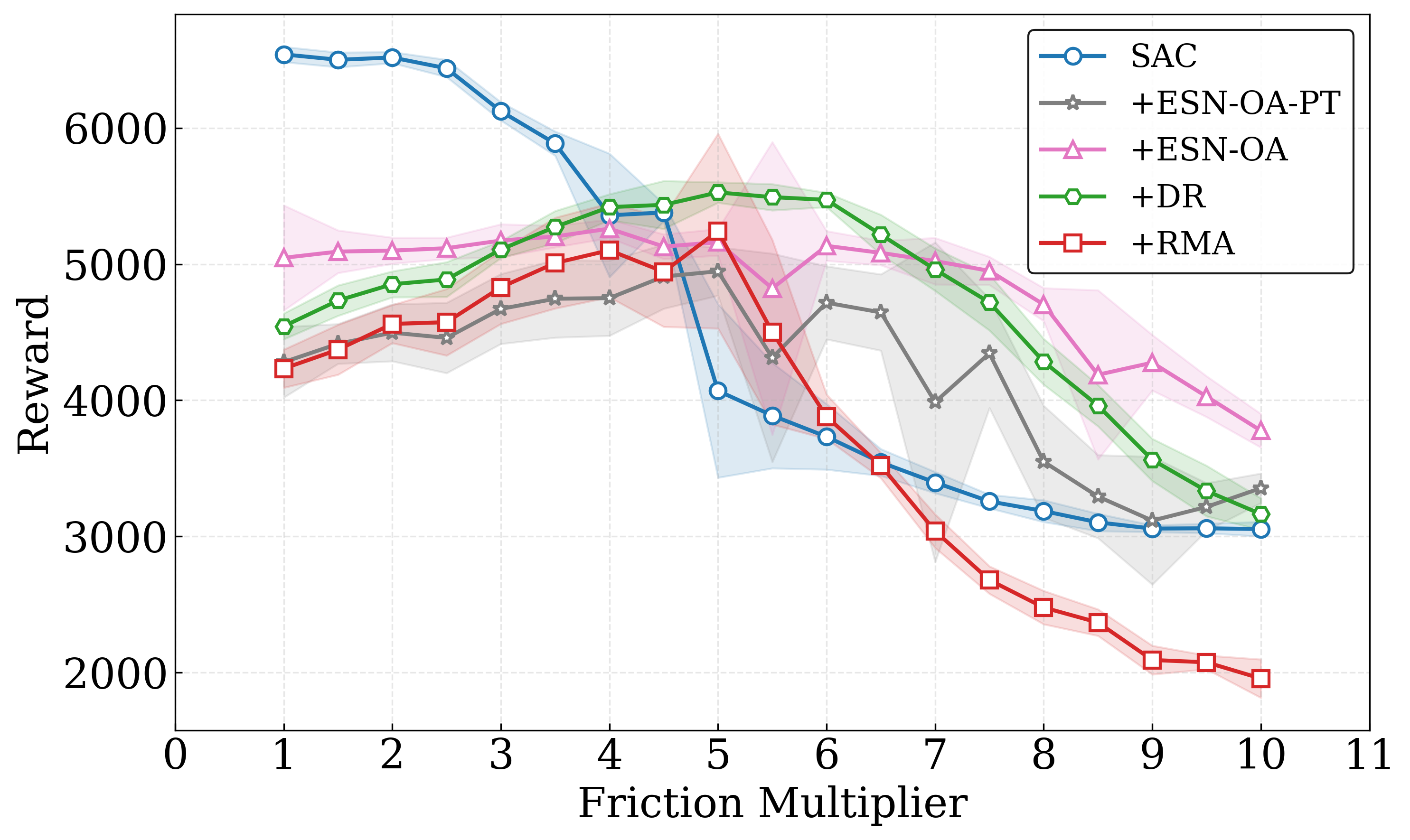}
    \caption{Average reward in the HalfCheetah environment under varying friction coefficients.}

    \label{fig:cheetah_result}
\end{figure}

\begin{table}[h]
\centering
\caption{Robustness evaluation on HalfCheetah (mean $\pm$ std).}
\label{tab:cheetah_stats}
\resizebox{\columnwidth}{!}{%
\begin{tabular}{l|r@{ $\pm$ }lr@{ $\pm$ }lr@{ $\pm$ }lr@{ $\pm$ }l}
\hline
\textbf{Method} & \multicolumn{2}{c}{\textbf{F=1}} & \multicolumn{2}{c}{\textbf{F=4}} & \multicolumn{2}{c}{\textbf{F=7}} & \multicolumn{2}{c}{\textbf{F=10}} \\ \hline
SAC & 6540.5 & 55.8 & 5358.4 & 453.7 & 3396.4 & 78.1 & 3054.1 & 53.6 \\
+ESN-OA-PT & 4279.8 & 259.1 & 4750.0 & 276.0 & 3986.5 & 1173.6 & 3354.8 & 108.4 \\
+ESN-OA & 5046.3 & 386.2 & 5260.6 & 68.2 & 5022.4 & 170.8 & 3777.2 & 123.4 \\

+DR & 4543.5 & 94.9 & 5418.7 & 95.4 & 4960.4 & 156.2 & 3164.3 & 116.6 \\ 
+RMA & 4232.8 & 140.0 & 5102.2 & 344.9 & 3039.3 & 123.2 & 1956.5 & 141.0 \\
\hline
\end{tabular}
}
\end{table}

Fig.~\ref{fig:cheetah_result} and Table~\ref{tab:cheetah_stats} summarize the results. The standard SAC agent achieved the highest performance at the nominal friction but degraded significantly as friction increased, indicating strong overfitting to the training dynamics. RMA and DR improved robustness in moderate regimes but collapsed in the extrapolation region. In contrast to the baselines, ESN-OA and ESN-OA-PT maintained stable performance across all friction settings. While ESN-OA achieved the highest reward under extreme conditions, ESN-OA-PT also demonstrated consistent robustness without the collapse seen in DR or RMA. These results demonstrate the effectiveness of the proposed online adaptation mechanism in handling severe and non-stationary dynamics shifts.

\subsubsection{Intra-Episode Adaptation Speed}
\begin{figure}
    \centering
    \includegraphics[width=0.79\textwidth]{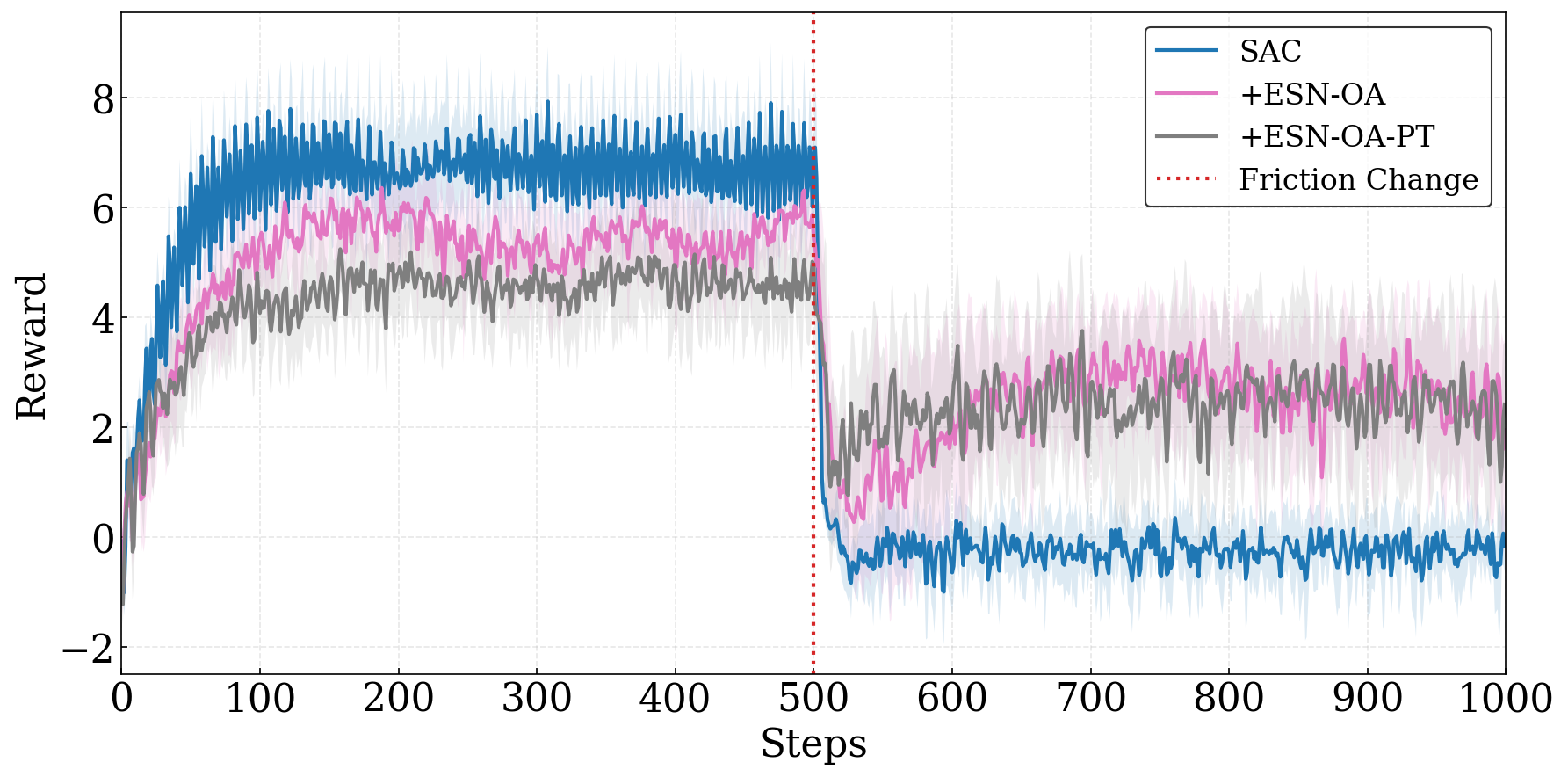}
    \caption{Reward transition after an abrupt friction change from $1.0\times$ to $10.0\times$ during an episode.}
    \label{fig:cheetah_step_reward}
\end{figure}

We analyzed the temporal adaptation process when the friction coefficient was abruptly switched from $1.0\times$ to $10.0\times$ during an episode. As shown in Fig.~\ref{fig:cheetah_step_reward}, the SAC agent failed immediately after the change, whereas the proposed ESN-OA rapidly recovered performance within a few dozen steps and maintained stable locomotion thereafter.

\subsubsection{ESN Output Weight Dynamics during Adaptation}

\begin{figure}[!b]
    \centering
    \includegraphics[width=0.75\textwidth]{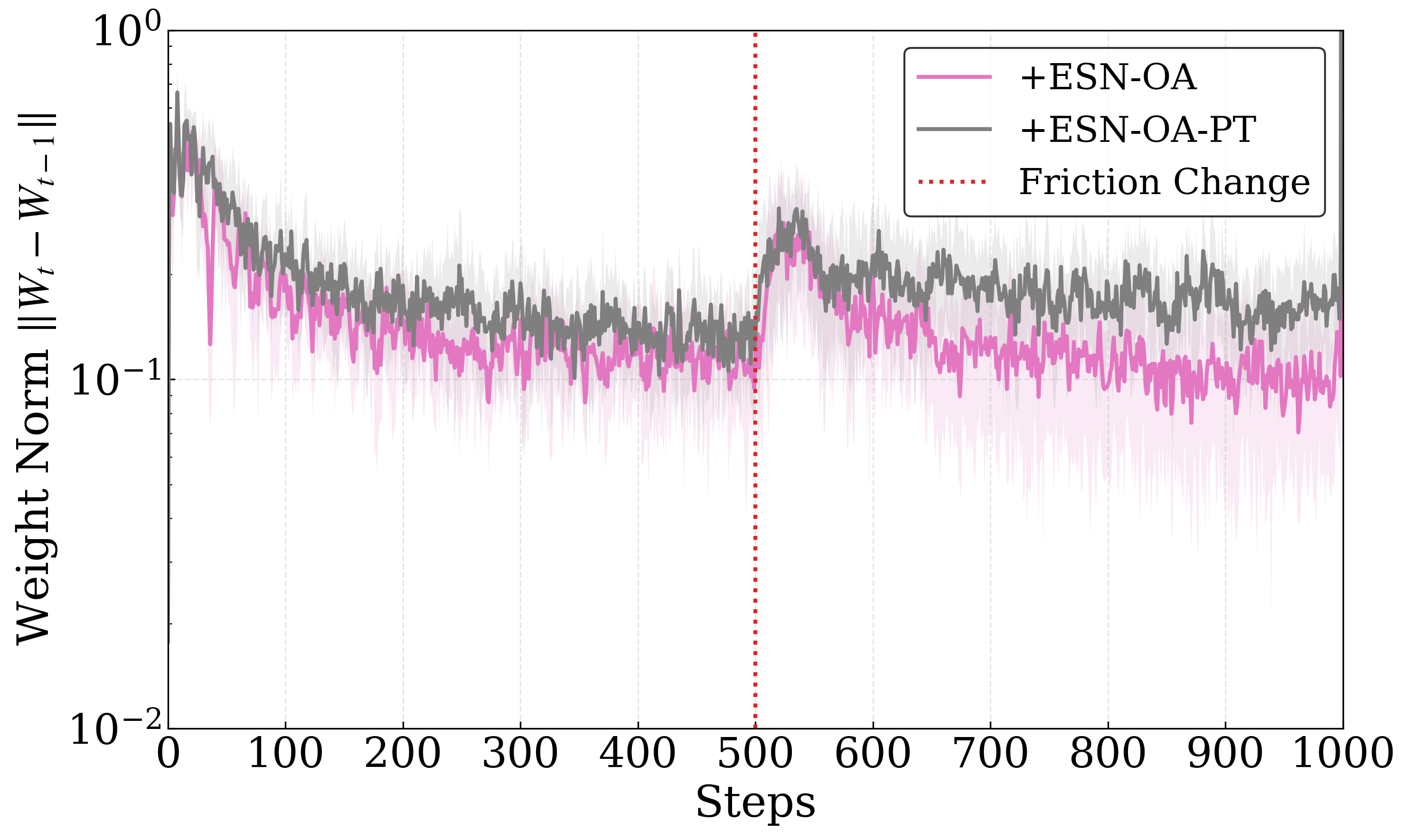}
    \caption{Temporal evolution of the ESN output weight update norm $\| W^{\mathrm{out}}_{t} - W^{\mathrm{out}}_{t-1} \|$ during friction change in HalfCheetah.}
    \label{fig:cheetah_weight_norm}
\end{figure}

To further analyze the adaptation behavior, we measured the temporal change in the ESN output weights.
Fig.~\ref{fig:cheetah_weight_norm} shows the norm of the step-wise difference
$\| W^{\mathrm{out}}_t - W^{\mathrm{out}}_{t-1} \|$
when the friction coefficient was abruptly changed during the episode. As shown in the figure, ESN-OA exhibits a pronounced spike in the output weight update norm immediately after the dynamics change, indicating rapid update of the readout parameters in response to large prediction errors.
In contrast, ESN-OA-PT shows more moderate updates, suggesting reduced sensitivity to sudden environment shifts.
These results highlight the fast and responsive nature of the proposed online adaptation mechanism.

\section{Discussion}

This work demonstrated that integrating the ESN-based online adaptation mechanism using the RLS online learning algorithm enables RL agents to rapidly adjust to non-stationary environments. Across both CartPole and HalfCheetah tasks, the proposed method consistently recovered performance within a few steps after abrupt changes in external disturbances or physical parameters, highlighting its effectiveness under severe OOD dynamics.
\subsection{Effectiveness of Online Adaptation}
The success of the proposed approach stems from two complementary properties. First, the reservoir dynamics provide a compact yet expressive temporal representation of recent observation histories, allowing the agent to implicitly infer latent environmental context that is not directly observable. This effectively addresses partial observability without relying on heavy recurrent architectures or long history windows. Second, the RLS-based readout adaptation achieves fast and stable convergence without backpropagation, making it suitable for real-time deployment and intra-episode adaptation.
\subsection{Comparison with Existing Methods}
Compared to DR, the proposed method does not depend on predefined parameter distributions and therefore remains effective even when test-time dynamics lie far outside the training distribution. In contrast to meta-RL approaches such as RMA, our framework requires neither privileged information nor extensive pretraining, significantly reducing deployment cost. Methods based on test-time training, such as PAD, can adapt online but incur higher computational overhead and may suffer from instability due to gradient-based updates. Restricting learning to a linear readout layer avoids catastrophic forgetting and ensures stable behavior during deployment.
\subsection{Intra-Episode Adaptation}
A key contribution of this study is the explicit evaluation of intra-episode adaptation, where environment dynamics change abruptly during task execution. The proposed method can continuously track such changes without resetting the policy or environment, which is critical for real-world robotic systems operating in non-stationary conditions.
\subsection{Limitations and Future Work}
Despite these advantages, the method has limitations. RLS can be sensitive to observation noise and outliers, and stability guarantees are not explicitly enforced. While suitable hyperparameter choices mitigated these issues in our experiments, incorporating formal safety constraints and adaptive regularization remains an important direction for future work.
Overall, the proposed ESN-based online adaptation framework offers a lightweight and practical alternative to existing robust and adaptive reinforcement learning methods. Its computational efficiency and ability to adapt without pretraining make it particularly promising for real-world and edge deployment in robotics and autonomous systems.


\vspace{12pt}

\bibliographystyle{IEEEtran}  
\bibliography{references}      
\end{document}